\documentclass{article}

\usepackage{arxiv}

\usepackage[utf8]{inputenc} % allow utf-8 input
\usepackage[T1]{fontenc}    % use 8-bit T1 fonts
\usepackage{hyperref}       % hyperlinks
\usepackage{url}            % simple URL typesetting
\usepackage{booktabs}       % professional-quality tables
\usepackage{amsfonts}       % blackboard math symbols
\usepackage{nicefrac}       % compact symbols for 1/2, etc.
\usepackage{microtype}      % microtypography
\usepackage{subcaption} 
\usepackage[hypcap=false]{caption}
\usepackage{float}
\usepackage{placeins}
\usepackage{flafter}
\usepackage{graphicx}
\usepackage{natbib}
\usepackage{doi}
 \usepackage{amsmath}
\title{A Large-Language-Model Assisted Automated Scale Bar Detection and Extraction Framework for Scanning Electron Microscopic Images}

\author{Yuxuan Chen\\
	Shanghai Jiao Tong Global College \\
    Shanghai Jiao Tong University\\
	Shanghai, 200240, China\\
	\texttt{cyx11020226@sjtu.edu.cn} \\ 
    \And
    Ruotong Yang\\
	Shanghai Jiao Tong Global College \\
    Shanghai Jiao Tong University\\
	Shanghai, 200240, China\\
	\texttt{ruotong2003@alumni.sjtu.edu.cn} \\ 
    \And
    Zhengyang Zhang\\
	Global Institute of Future Technology \\
    Shanghai Jiao Tong University\\
	Shanghai, 200240, China\\
    \texttt{zhengyangzhang@sjtu.edu.cn} \\
    AI Lab, Xiaomi Corporation, China \\
	\texttt{zhangzhengyang3@xiaomi.com} \\
    \And
 \href{https://orcid.org/0000-0003-0996-3156}{\includegraphics[scale=0.06]{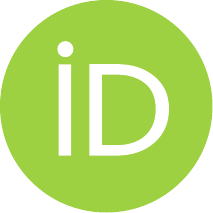}\hspace{1mm}Mehreen Ahmed}\\
 Global Institute of Future Technology \\
    Shanghai Jiao Tong University\\
	Shanghai, 200240, China\\
   \texttt{mehreen@sjtu.edu.cn} \\
	\And
	\href{https://orcid.org/0000-0002-0912-681X}{\includegraphics[scale=0.06]{orcid.pdf}\hspace{1mm}Yanming Wang} \\
	Global Institute of Future Technology \\
    Shanghai Jiao Tong University\\
	Shanghai, 200240, China\\
 \texttt{yanming.wang@sjtu.edu.cn} \\
}

\begin{document}
\maketitle

\begin{abstract}
Microscopic characterizations, such as Scanning Electron Microscopy (SEM), are widely used in scientific research for visualizing and analyzing microstructures. Determining the scale bars is an important first step of accurate SEM analysis; however, currently, it mainly relies on manual operations, which is both time-consuming and prone to errors. To address this issue, we propose a multi-modal and automated scale bar detection and extraction framework that provides concurrent object detection, text detection and text recognition with a Large Language Model (LLM) agent. The proposed framework operates in four phases; i) Automatic Dataset Generation (Auto-DG) model to synthesize a diverse dataset of SEM images ensuring robust training and high generalizability of the model, ii) scale bar object detection, iii) information extraction using a hybrid Optical Character Recognition (OCR) system with DenseNet and Convolutional Recurrent Neural Network (CRNN) based algorithms, iv) an LLM agent to analyze and verify accuracy of the results. The proposed model demonstrates a strong performance in object detection and accurate localization with a precision of 100\%, recall of 95.8\%, and a mean Average Precision (mAP) of 99.2\% at IoU=0.5 and 69.1\% at IoU=0.5:0.95. The hybrid OCR system achieved 89\% precision, 65\% recall, and a 75\% F1 score on the Auto-DG dataset, significantly outperforming several mainstream standalone engines, highlighting its reliability for scientific image analysis. The LLM is introduced as a reasoning engine as well as an intelligent assistant that suggests follow-up steps and verifies the results. This automated method powered by an LLM agent significantly enhances the efficiency and accuracy of scale bar detection and extraction in SEM images, providing a valuable tool for microscopic analysis and advancing the field of scientific imaging.
\end{abstract}

% keywords can be removed

\section{Introduction}
\label{sec1:into}
Currently, laboratories and production lines around the world are constantly generating a large amount of image data with scale bars, using various devices such as cameras, optical microscopes, electron microscopes, and so on \citep{ref1,ref2}. These scientific images include an essential scale information that is represented by a bar (of a known length) to help the viewers relate the image to the actual object. This scale information provides the quantitative evaluation or a visual reference i.e., the fiber diameter, particle size etc. that allows researchers to determine the size and scale of the observed structures in the fields of materials and medicine. In the realm of scientific and industrial exploration, microscopic images play a crucial role in visualizing and analyzing micro structures. These high-resolution images provide researchers with valuable insights into the intricate details of various materials and biological specimens. Accurate interpretation and analysis of Scanning Electron Microscopy (SEM) images require precise information of the magnification level and physical dimensions of the objects within the image, i.e., a process for mapping the size of image to real world dimensions. To address this need, scale bars are commonly included in SEM images. However, the detection and extraction of scale bars have traditionally been manual tasks, relying on human operators to identify and measure the scale bars within the images. This manual approach is not only time-consuming and labor-intensive but also susceptible to human error, leading to inaccuracies in the subsequent analysis \citep{ref3,ref4}. The significant diversity in the scale patterns, low contrast and blurry vision can make the manual detection more complex than previously anticipated \citep{ref5}. Therefore, the automated identification of scale information whether it is for data recording of a single production line or for forming datasets of similar images, automatically processing and analyzing image data to significantly improve processing efficiency and accuracy is a common expectation in both academia and industry. 

Furthermore, with the gradual application of Artificial Intelligence (AI) algorithms and automated equipment to various domains, automating the processing and analysis of image data is also a necessary step to achieve self-driven laboratories and industrial intelligence. Prior methods for scale bar detection have primarily relied on classical image processing tools or semi-automated approaches with significant limitations. ImageJ and its Fiji distribution \citep{ref6,ref7}, while widely used in microscopy analysis, require manual input of scale information or depend on embedded metadata that is often incomplete or inconsistently formatted across different microscope systems. OpenCV-based solutions attempt to automate detection through edge detection and template matching \citep{ref8}, or rely on geometric and color-based heuristics, such as enforcing high aspect ratios and intensity thresholds, to identify scale bars \citep{ref9}. However, these methods fail when dealing with variable scale bar designs, low-contrast conditions, or complex backgrounds. 

With the rise of deep learning technology, the automatic identification of image scales can be achieved using object detection and Optical Character Recognition (OCR) techniques \citep{ref10}. Many researchers have developed various methods and tools to automatically extract quantitative information from images and achieved significant results in their respective fields \citep{ref11}. For example, a study \citep{ref12} integrated advanced image processing, robust preprocessing like noise reduction, background correction, and precise segmentation/feature extraction with diverse microscopy techniques including light, fluorescence, super-resolution, and electron to revolutionize cellular biology. Furthermore, these studies emphasize the growing importance of AI and multi-modal imaging for in depth analysis, and opening new frontiers for discovery through interdisciplinary innovation. Another study \citep{ref13} developed a universal deep learning framework to perform fast and accurate online segmentation, shape extraction, and statistical analysis of nanoparticle morphology in complex SEM images and achieved a classification accuracy of 86.2\%. However, handling non-standard scale bars (e.g., colored or gradient-filled) which underscores the need for a more adaptive, context-aware solution is still lacking. 

The recent advancements in Large Language Models (LLMs) architecture have significantly improved their capacity to handle complex tasks and offer a new paradigm by providing reasoning over multi-modal outputs (e.g., detected bounding boxes, OCR text, and spatial or colored layouts) \citep{ref14, ref15}. These LLMs allow users to ask questions in natural language without any technical terms related to SEM imagery and can intelligently interpret context, resolve ambiguities, and make domain-informed decisions by providing a context-aware scientific interpretation. This capability makes them uniquely suited to overcome the challenges of scale bar interpretation in complex or diverse microscopy imagery and provides opportunities for automating complex scientific workflows in a more adaptive and scalable manner. A study \citep{ref16} introduced a two-step AI-driven framework that combines a deep learning model for defect classification with a LLM based assistant for real-time, user-friendly guidance that is trained on a curated defect Atomic Force Microscopy (AFM) dataset and achieves high accuracy in detecting key defects. A framework is proposed in \citep{ref17} that automates AFM using LLM agents by highlighting the critical need for improved benchmarking and prompt engineering before reliable deployment in self-driving labs.

In this study, we propose a novel AI-driven framework that automates key steps in materials characterization from SEM images. The framework leverages deep learning techniques, including a You Only Look Once (YOLO)-based object detection model and a hybrid OCR algorithm for text recognition. It utilizes a dataset curated by an Automatic Dataset Generation (Auto-DG) module and integrates computer vision with LLM reasoning to act as a scientific agent that analyzes, validates, and contextualizes the extracted information. The Auto-DG helps in comprehensive training, by locating and extracting scale bars from SEM images with high precision and reliability. The proposed system offers a unified approach that can simultaneously perform object detection, text recognition and detection for scale bar detection in SEM images using a hybrid OCR approach with a dual-engine strategy employing Chinese OCR (CnOCR) for general text recognition, Parallel Distributed Deep Learning OCR (PaddleOCR) for numeral-specific tasks, and Euclidean distance-based association of text with scale bars. The development of an automated scale bar detection model has the potential to revolutionize the analysis of SEM images, enabling researchers to streamline their workflows, reduce human error, and unlock new possibilities for scientific exploration. In this regard, the proposed model can efficiently and precisely analyze the scale information within a large-scale SEM image database. The LLM based hybrid approach enables high-level interpretation such as verifying measurement consistency, detecting anomalies, and suggesting next steps in analysis. The key advantage of this proposed method is that it enables the automation of the scale bar extraction process, eliminating the need for manual measurement and the associated risks of human error. This automated model can help researchers streamline their workflows and gain access to accurate scale information directly from the SEM images, even when dealing with a vast repository of data offering a practical and extensible path toward intelligent, interpretable, and scalable scientific image analysis.

\section{Methodology}\label{sec2:method}
The proposed framework is a multi-stage system designed to automatically detect and extract scale bars from SEM images addressing the complex graphical and semantic challenges of scale bar detection, creating an automated data generation process to tackle data scarcity and diversity, and enhancing the model’s generalization capabilities across various scenarios. It combines the strengths of object detection, OCR, synthetic data generation, and LLM reasoning. The architecture of the proposed system, as detailed in Figure \ref{fig1}, encompasses a four-tiered methodology: (i) an Auto-DG module that creates a diverse, labeled dataset of SEM images with synthetic scale bars to ensure robust model training and generalization; (ii) an object detection module, which employs deep learning and object detection networks to localize scale bars in the images; (iii) an information extraction module with a hybrid OCR system for robust scale value and unit extraction; and (iv) a LLM agent that performs verification, reasoning, and interactive feedback to refine and validate the extracted results. Initially, an Auto-DG model is used to generate a diverse dataset for the nuanced training of the proposed framework to support the robustness and generalization of the model. This diverse dataset of SEM images with manually annotated scale bars ensures the model can handle various scale bar formats and image conditions that is crucial for training the detection model and enhancing its performance across different scenarios. Subsequently, the YOLO-based architecture is deployed, leveraging the dataset curated by the Auto-DG, where it identifies the presence and precise positioning of scale bars within the images. This stage employs a deep learning based image detection model to accurately locate scale bars, regardless of their position or orientation. Following object detection, the system proceeds to text recognition. It identifies and extracts text information associated with the detected scale bars. By analyzing the proximity of text regions to the scale bar, the system selects the text with the minimum distance as the scale value. This text, typically indicating the scale (e.g., ``XX$\mu$m''), is then processed for further analysis. The extraction process is refined to transmute the textual data into a mathematical representation, thereby facilitating rigorous analysis and measurements predicated on the foundational scale bar information. This modular design allows the system to address key challenges in scale bar analysis, including varied visual styles, noisy image backgrounds, and OCR ambiguities. Once the system has detected the scale bar and its value, it is passed to an LLM agent to analyze and verify the results. A user interface is provided to visualize the generated report by the LLM agent. The system prompt to initialize the LLM conversation and define the model’s role are given in Figure \ref{fig1}. The LLM agent first gives a general analysis on the performance of the proposed model by commenting on the output and suggests any improvements needed. Next, the LLM can also respond to the user queries and give its feedback.

\subsection{Dataset Generation and Training}
The data used for automatic dataset generation is acquired from two different sources (i) NFFA-EUROPE - Majority SEM Dataset \citep{ref18} and (ii) Image Data Resource (IDR) dataset \citep{ref19}. The NFFA-EUROPE dataset is the first annotated collection of high-resolution 21,272 SEM images that are classified into 10 categories focused on nanomaterials and nanostructures, such as; i) porous sponge, ii) patterned surface, iii) particles, iv) films and coated surfaces, v) powder, vi) tips, vii) nanowires, viii) biological, ix) MEMS devices and electrodes, x) fibers. Each category contains 1,024 × 728 pixels jpg files assembled according to the majority criterion \citep{ref20}. The IDR dataset is a public resource that has biological imaging datasets from 24 imaging studies on diverse biological contexts such as genome-wide RNAi screens, developmental biology, and infectious disease on 20+ model organisms e.g., human, mouse, fly, plant and fungal cells, marine organisms etc. are included with $~$50 TB of image data. 

No open-source datasets are available for scale bar detection and extraction; thus a diverse scale bars dataset is simulated with these provided SEM images taken from NFFA dataset. The SEM images in this dataset have scale bars in fixed locations, therefore, the images are preprocessed by simply cropping the scale bars. To address the challenge of manually annotating scale bar positions in real-world datasets for training purposes, a novel approach is developed to automate the dataset generation process. Specifically, this model is designed that automatically synthesizes training data by programmatically augmenting SEM images with scale bars. This method creates a diverse set of synthetic images with scale bars of varying shapes, sizes, and orientations, thereby significantly enhancing the generalization of the proposed model. During the generation process, the parameters of each scale bar, including its coordinates, height and width is recorded which are essential for training object detection models. By leveraging this automated dataset generation approach, a large and varied dataset is produced that supports robust model training without requiring extensive manual annotation. 

Figure \ref{fig2}a gives an overview of the data generation and training process. The process begins by manually drawing common shapes of scale bars such as straight lines, line segments, ruler shapes, and truncated shapes with text in the middle. These common shapes are categorized as follows: i) joint-label bar, ii) I-shaped bar, iii) ruler-shaped bar and iv) rectangular bar, as seen in Figure \ref{fig2}b. These manually created scale bars are then combined with an existing SEM image dataset obtained from \citep{ref18}, consisting of a collection of SEM images. The original scale bars present in the SEM images are first removed by cropping out the region containing the scale bar, retaining only the microscopic image part as the background for scale dataset generation. Next, these annotated SEM images with the manually drawn scale bars are fed into the Auto-DG model. This process systematically generated a large and varied dataset with annotated scale bar positions and lengths, significantly enhancing the model’s ability to generalize across different scenarios. During the scale bar generation process, parameters such as position and size of each scale bar are automatically determined and stored. Each file corresponds to a SEM image and contains information about the labels, bounding box coordinates, and categories of each scale bar.

Additionally, random textual information, such as ``100 mm'', is generated and annotated below the scale bars. This approach resulted in a comprehensive dataset that is divided into training, validation, and testing sets, facilitating the model’s subsequent training, evaluation, and testing. By following this dataset generation process, a dataset comprising four common shapes of scale bars and randomly generated textual information is successfully created. This dataset can be employed for training object detection models to accurately detect and localize scale bars in SEM images while recognizing the textual information below the scale bars.

\subsection{Object Detection}
Further advancements in image detection methodologies have been made possible by utilization of deep learning techniques. For example, the integration of Convolutional Neural Networks (CNNs) with object detection models has been a significant area of research, leading to the development of models that can more effectively capture and analyze spatial hierarchies in images. These methodologies have expanded the possibilities for image detection, allowing for more sophisticated and nuanced object recognition capabilities. In the object detection phase, the YOLOv5 deep learning architecture \citep{ref21} is employed. The training process starts with the diverse dataset (provided in Appendix A, Figure A1), which is meticulously generated using the Auto-DG framework as described previously. YOLOv5 has played a key role in accelerating image detection research by offering a speedy, precise and an efficient means of detecting objects within images. YOLOv5 significantly improves upon the accuracy and speed of object detection by adopting features like automatic batch size and learning rate adjustment, improved model scaling, and enhanced data augmentation techniques. This makes it highly adaptable for a variety of applications, from surveillance systems to autonomous driving technologies. For SEM scale bar detection task, this model proved to be robust and impactful as compared to other object detection models.

% Figure
\noindent
\begin{center}
\includegraphics[width=1\textwidth]{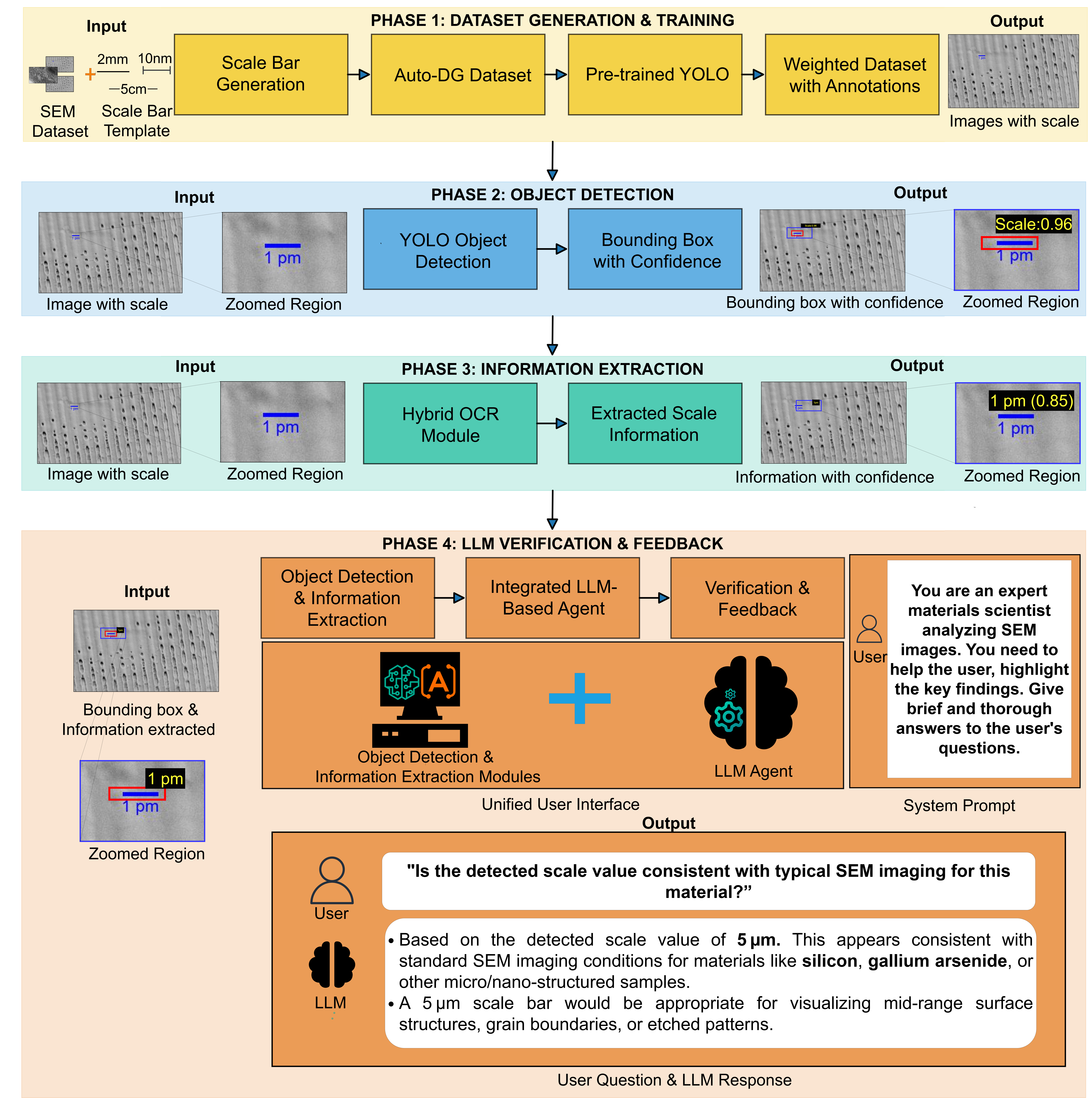}
  \captionof{figure}{High level architecture outlining the multi-stage methodology, i.e., Phase i: Dataset Generation and Training, Phase ii: Scale Bar Object Detection, Phase iii: Information Extraction and lastly, Phase iv: LLM Verification and Feedback, for our proposed system that is designed for SEM images.}\label{fig1}
\end{center}

Using the YOLO model, the system learned to detect scale bars by optimizing its parameters through back-propagation and gradient descent. This involved minimizing a loss function that measures the discrepancy between the predicted bounding boxes and the ground truth annotations. The annotated Auto-DG dataset is used to train the model, ensuring exposure to a wide variety of synthetic SEM images with diverse backgrounds, aspect ratios, and bar designs. Next, training is performed over multiple epochs with image augmentation techniques such as mosaic, random cropping, and brightness jittering to improve generalization. The final trained model yields a ``best.pt'' file representing the optimal weights and parameters that achieved the highest mAP@0.5 on the validation set. During inference, the YOLO model outputs bounding boxes and associated confidence scores for the detected scale bars, which are then passed to the OCR module for further analysis.
% Figure

\begin{center}
\includegraphics[width=1\textwidth]{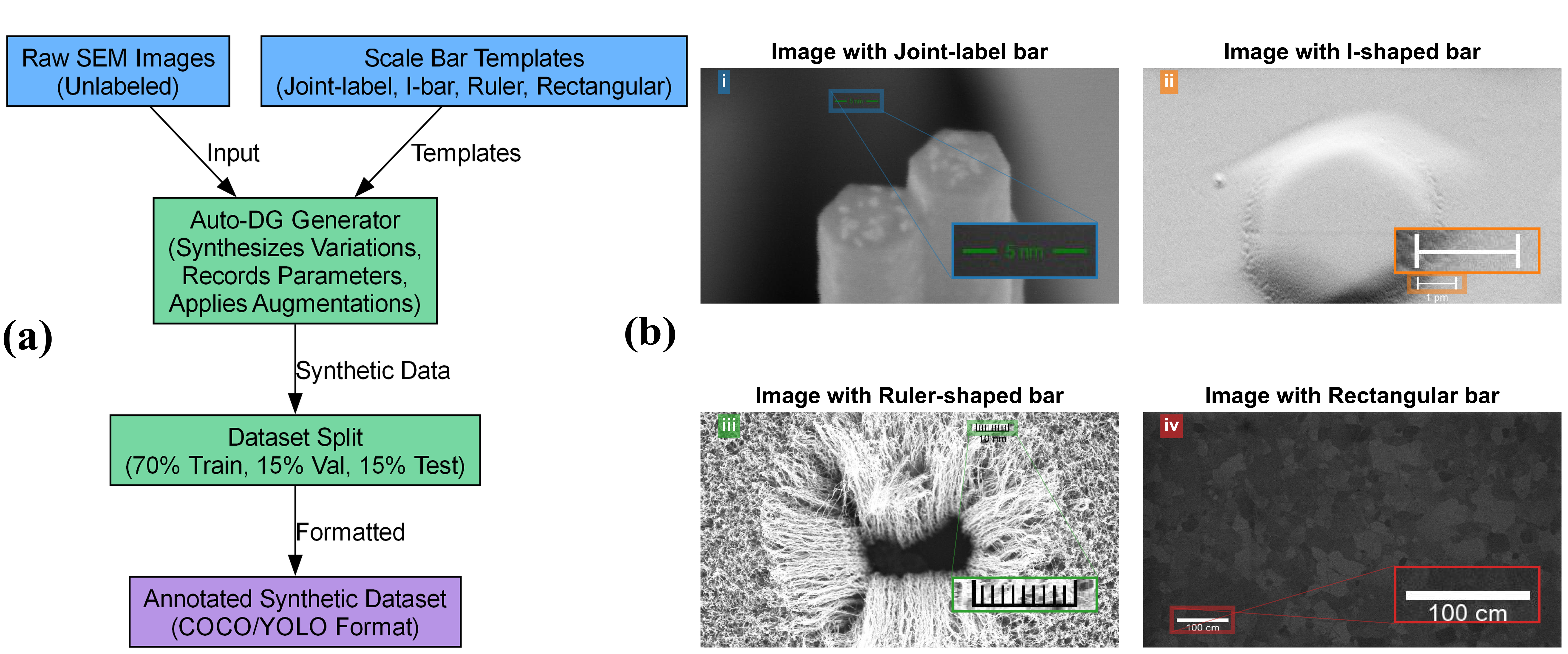}
\captionof{figure}{(a) Flowchart outlining the data generation and training process. (b) Four common shapes of scale bars: i) Joint-label bar, ii) I-shaped bar, iii) Ruler-shaped bar and iv) Rectangular bar.}\label{fig2}
\end{center}
\subsection{Information Extraction}
Once the scale bar is detected, the system proceeds to the information extraction phase. In this phase, the CnOCR \citep{ref22} model that is developed for highly efficient and accurate Chinese text detection while handling a wide range of text recognition tasks from scanned documents to natural scene images, is used. The tool supports multiple languages, making it versatile for global applications. It's design emphasizes ease of use and flexibility, offering pre-trained models that can be quickly deployed or fine-tuned for specific needs. The lightweight design of this OCR tool ensures that it can be run on various platforms, facilitating the development of OCR applications in areas like automated document processing, intelligent navigation systems, and educational tools. Another important OCR tool, PaddleOCR \citep{ref23}, developed by Baidu, is a robust solution for multi-language OCR tasks. It supports numerous languages and is built on cutting-edge models that excel in general text recognition tasks. However, as observed in the experimental results, PaddleOCR may exhibit lower precision in recognizing numerals, particularly in contexts where numerals are embedded in complex backgrounds \citep{ref24, ref25}.

Therefore, in this phase, the coordinates of the center point of the scale bar detection box are retained and then PaddleOCR is employed to process the image to extract the associated text. Firstly, the text box is found with specific measurement units such as, `$cm$', `$mm$', `$\mu$m', `$nm$' and `$pm$' and then the location of box center is stored. To ensure a robust search for units like `$\mu$m' that may also be represented as the `um', the proposed methodology incorporates a dual-check mechanism, thereby circumventing potential detection pitfalls. The ensuing step entails a calculated assessment of the proximity between the scale bar and each valid text box, ultimately identifying the nearest one by the minimal distance (${D}_{min}$) as shown in Equation \ref{eq1}.

\begin{gather}\label{eq1}
\bar{X}_{\mathrm{value}}, \bar{Y}_{\mathrm{value}} = \arg\min_{X_{i,\mathrm{text}}, Y_{i,\mathrm{text}}} \sqrt{ (\bar{X} - X_{i,\mathrm{text}})^2 + (\bar{Y} - Y_{i,\mathrm{text}})^2 }
\end{gather}

Here, the $\bar{X}$, $\bar{Y}$ is the center point of the bar detection box, and the $X_{i,\mathrm{text}}$, $X_{i,\mathrm{text}}$ is the center point of the text detection box. The text within this optimally located box is presumed to represent the scale bar’s value. After extracting the text, it is translated into a mathematical representation to obtain the real length or area of the objects in the SEM image. These advancements significantly contribute to the efficiency and accuracy of scale bar detection in SEM images, providing a valuable tool for microscopic analysis and scientific research.

\subsection{LLM Verification and Feedback}
The LLM agent serves as a context-aware reasoning module that operates after the object detection and information extraction phases to validate, refine, and guide the interpretation of results. This proposed LLM agent has the object detection and information extraction context fed into the backend, allowing it to reason using structured data like pixel length, OCR value, unit, confidences, etc. before generating its answer from the SEM image analysis pipeline. First, the object detection and information extraction modules are run to detect scale bars and extract the scale value and units from the SEM image. Then, the LLM agent (e.g., LLaMA3-70B with LangChain) is used to analyze the results in the broader semantic and spatial context of the image, ensuring that the extracted values are logical, consistent, and aligned with standard conventions in SEM imagery. The prompt is designed as structured inputs i.e.; bounding box coordinates, the OCR output and detection confidence scores are formatted into natural language prompts. These prompts enable the LLM to assess whether the detected scale bar values match expected patterns (e.g., ``5 $\mu$m'' or ``200 $nm$''), flag potential mismatches or anomalies (e.g., numeric-text inconsistencies), and suggest corrections or follow-up steps. For instance, if the OCR detects ``50 $\mu$m'' and the context suggests a different scale, the LLM can propose re-evaluation or alert the user. Additionally, the agent can recommend adjustments to the pipeline, such as retraining on specific failure cases or applying a correction filter. These interactions make the LLM a dynamic assistant, not only verifying outputs but also adapting the pipeline in an intelligent, user-informed manner.

\section{Experiments and Results}\label{sec3:experiments}
The proposed model is extensively tested to ensure its generalization and effectiveness across a broad range of scenarios, including scale bars located in varying positions, against diverse background colors, and on additional backgrounds with distracting text common in many scientific publications. The results demonstrate the model’s robustness in detecting and extracting scale bars regardless of the visual complexity of the images, which is critical for researchers requiring precise measurements from SEM images where the scale bar may not be prominently displayed or might blend subtly into the background. The proposed method is much faster and performs well for small object detection problems as compared to other object detection techniques for scale bar detection \citep{ref26, ref27}. Object detection models like YOLOLite, NanoDet take longer training time and giver poor results. For scale bar information extraction, other popular OCR techniques including; EasyOCR, CnOCR and PaddleOCR were also explored. 

Figure \ref{fig3}a, gives the architecture flow of the object detection process for the scale bar detection. Figure \ref{fig3}b shows the outputs of the object detection model on the Auto-DG dataset. As seen, the image displays the zoomed region of the five separate scenarios where the scale bars are positioned differently i.e.; top-right, top-left, bottom-left, bottom-right and center, demonstrating the proposed model’s capability to recognize scale bars regardless of their orientation or position on the SEM image. This is essential for ensuring accurate spatial measurements in varied scientific imaging contexts and more challenging conditions. It features scale bars placed against highly textured and dark backgrounds showcasing the model’s robustness and high level of precision in complex visual environments. These images highlight the model’s advanced processing abilities to filter and focus on relevant scale bar information despite potential visual interference. 

This rigorous testing ensures that the proposed model remains a reliable tool for researchers dealing with a wide range of SEM imaging conditions. The final evaluation of the object detection model offers improved accuracy and faster inference speeds with a higher precision (100\%) and recall (95.8\%) in detecting small objects like scale bars, due to better feature extraction and enhanced anchor box strategies. Additionally, the lightweight design of this architecture makes it easy for deployment on various devices while maintaining robust performance.

Figure \ref{fig4}a presents the architecture flow of the information extraction phase. For information extraction, a hybrid OCR approach was developed and evaluated for accurate detection of scale bar values in SEM images, demonstrating significant improvements over standalone OCR engines. This hybrid approach leverages a dual-engine strategy employing CnOCR for general text recognition and PaddleOCR for numeral-specific tasks, the method integrates enhanced post-processing steps, including unit normalization (e.g., $\mu$m → um), strict word-boundary checks to eliminate false positives (e.g., distinguishing``um''in ``summary''), and Euclidean distance-based association of text with scale bars. The outputs of the information extraction process with the hybrid CnOCR and PaddleOCR algorithms can be observed in Figure \ref{fig4}b. The system successfully identifies scale bar values such as ``10 cm'' and ``500 mm'' with high confidence scores (ranging from 0.53 to 0.92). Notably, the system maintains robust performance across multiple instances, as evidenced by repeated detections of ``5 $\mu$m'' with stable confidence scores (0.53–0.54). The lower confidence scores (e.g., 0.53 – 0.54) may indicate challenges with blurred or low-contrast text, suggesting areas for further refinement. Overall, these results highlight the hybrid approach's precision, adaptability, and ability to distinguish relevant scale bar information from background noise, making it a reliable tool for automated metrology in SEM imaging. The image results showcase the hybrid OCR system’s robust performance across diverse SEM scale bar configurations, with confidence scores reflecting its adaptability to varying conditions.
% Figure

\begin{center}
\includegraphics[width=1\textwidth]{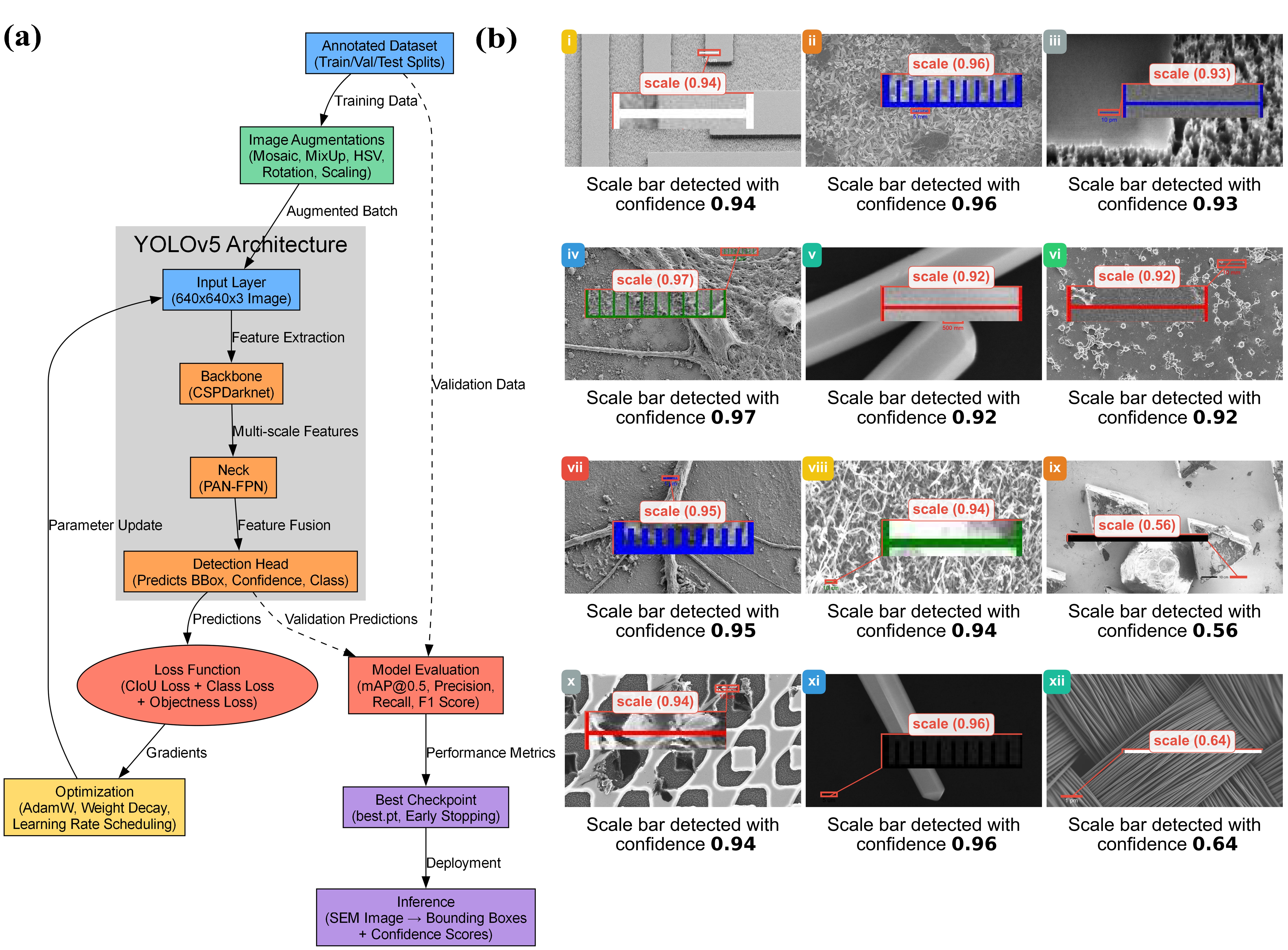}
\captionof{figure}{(a) Flowchart outlining the architecture of the object detection phase. (b) Performance of the proposed object detection model in detecting scale bars is demonstrated using SEM images with complex backgrounds and scale bars placed in different positions i.e.; top-right (i, iv, vi, x), bottom-left (viii, xi, xii), bottom-right (ix), center (ii, iii, v, vii). The detected scale bars are highlighted with red bounding boxes and confidence scores in parenthesis and zoomed-in regions are shown in the center. }\label{fig3}
\end{center}

The evaluation on the Auto-DG dataset, the combined OCR approach achieved 89\% precision and a balanced 75\% F1 score , outperforming individual engines such as EasyOCR (58\% F1), CnOCR (47\% F1), and PaddleOCR (41\% F1), as seen in Figure \ref{fig4}c. Notably, the system reduced false positives as compared with EasyOCR and CnOCR while maintaining 65\% recall, prioritizing reliability over exhaustive detection. This is a critical advantage for scientific applications where erroneous scale interpretations could compromise data integrity. While the method exhibits a precision-recall trade-off, its high F1 score and domain-specific optimizations make it a superior solution for SEM image analysis, offering a scalable, automated alternative to manual verification. The success of this approach lies in its context-aware processing, robust unit validation, and intelligent fallback mechanism, making it a reliable solution for automated scale bar detection in scientific imaging. Future refinements could further optimize recall, but the current implementation already sets a high benchmark for precision and robustness in OCR-based metrology. 

The object detection and information extraction models are also evaluated on the same test set, consisting of 25 real-world images from our laboratory to ensure realistic testing conditions (See Appendix A, Figure A2, A3). Figure \ref{fig5} shows the outputs of the proposed object detection model on these test images. As seen in Figure \ref{fig5}, an accurate scale bar detection even with scale bars located in varying positions, against diverse background colors, and distracting text common in many scientific publications. In addition, the results show the model’s robustness in detecting and extracting scale bars regardless of the visual complexity of the images, which is critical for researchers requiring precise measurements from SEM images where the scale bar may not be prominently displayed or might blend subtly into the background.

While the proposed framework has shown promising results in detecting and extracting scale bars from a variety of SEM images, it is not without limitations. One notable challenge encountered during the experiments is the model’s performance in colorful images. The current version of the framework tends to exhibit a lower recognition accuracy when dealing with scale bars set against vibrant or multicolored backgrounds. This limitation can be attributed to the model’s training primarily on grayscale SEM images, where the contrast between the scale bar and the background is typically high. Furthermore, enhancing recognition efficiency remains a pivotal goal. Optimizing the model’s architecture to reduce computational demands while maintaining or improving accuracy can lead to faster processing times, making the proposed system more practical for real-time applications. Implementing these improvements will not only extend the model’s applicability but also ensure that it remains a valuable tool in the rapidly advancing field of microscopic image analysis.

% Figure
\begin{center}
\includegraphics[width=1\textwidth]{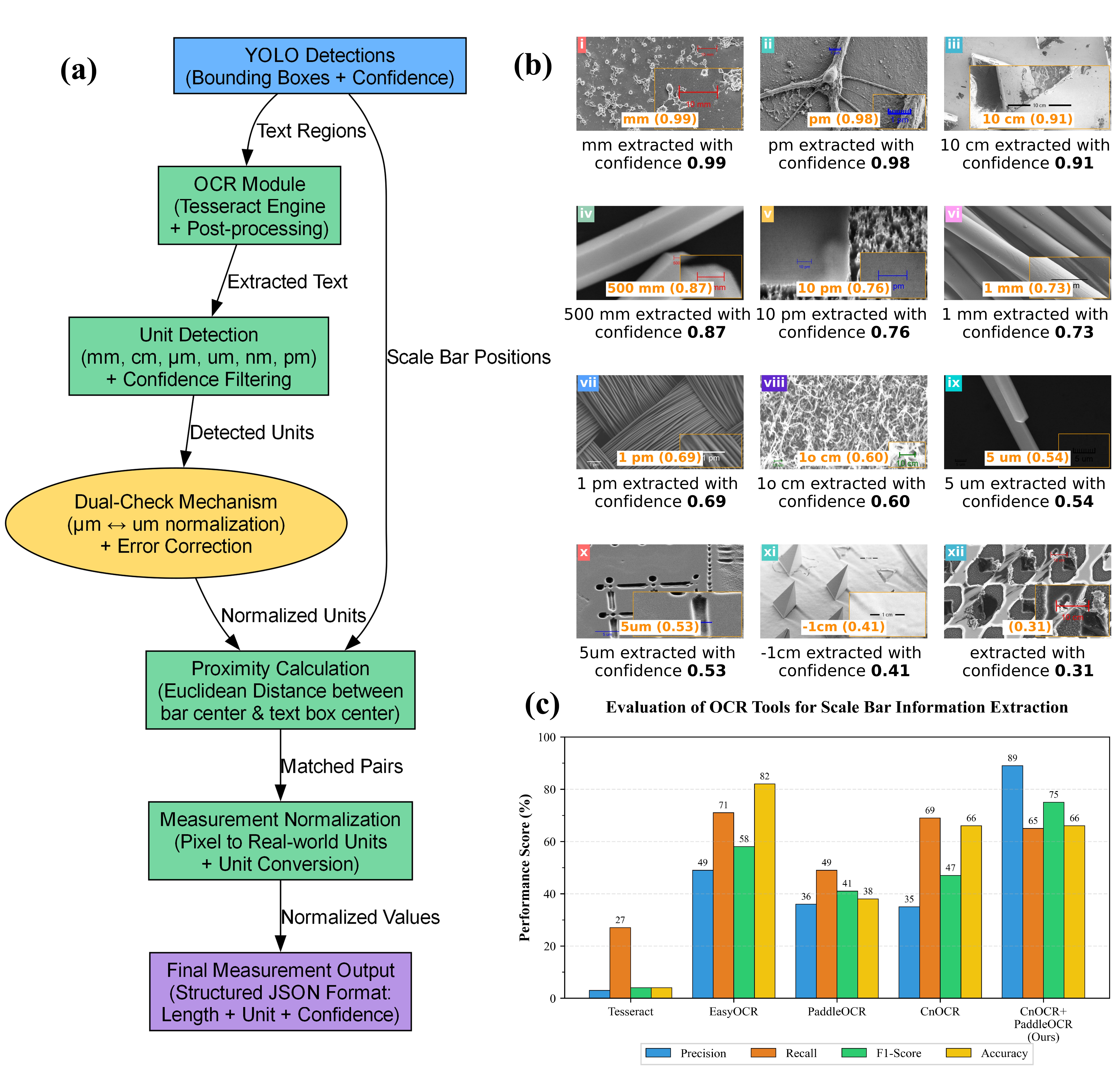}
   \captionof{figure}{(a) Flowchart outlining the steps involved in the information extraction phase. (b) The example outputs from the hybrid OCR system for SEM scale information extraction are shown that detected values (e.g., ``10 cm'', ``500 mm'') are displayed with confidence scores (in parenthesis). The system handles variations in nominal values (e.g., iv: 87\%, vi: 73\%), units (e.g., iii: 91\%, v: 76\%), and scale mismatches (e.g., vi: 73\%, vii: 69\%), demonstrating robustness across diverse conditions. (c) The performance evaluation of different OCR tools for the proposed information extraction on the SEM images.}\label{fig4}
\end{center}
% Figure
\begin{center}
\includegraphics[width=0.8\textwidth]{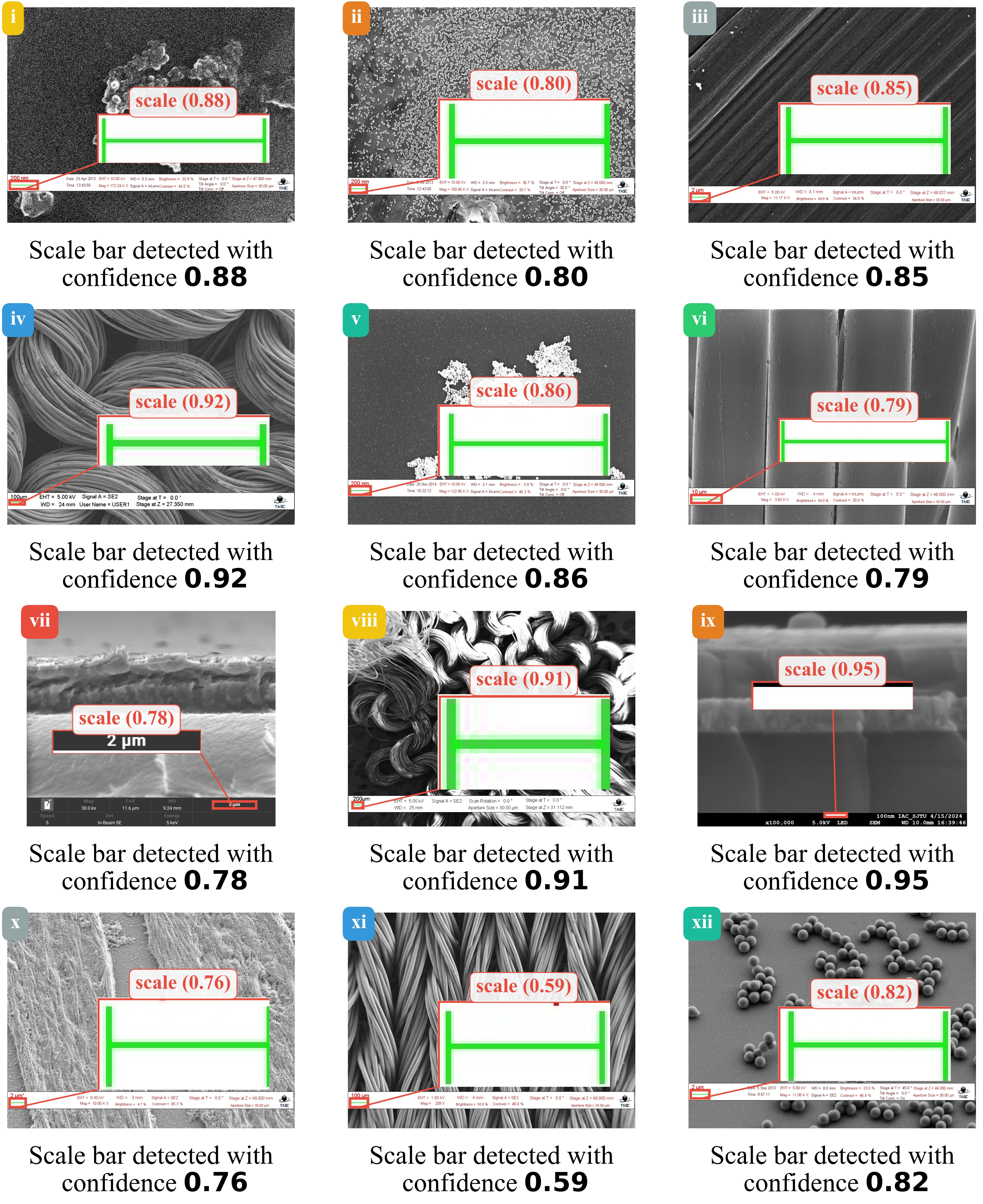}
   \captionof{figure}{Proposed framework performance on real-world SEM images from our laboratory, with detected scale bars highlighted by bounding boxes and confidence scores (zoomed regions shown).}\label{fig5}
\end{center}

However, to overcome such limitations, we have proposed the LLM assistance and guidance to help improve experimental conditions and modify techniques for scale bar detection and text extraction from SEM images. This interactive approach contributes in more efficient and reliable SEM image analysis. The proposed system includes a user-friendly web interface that allows researchers to upload SEM images and visualize the automated scale bar detection and extraction results in real time (See Appendix A, Figure A3). This web interface displays the detected scale bar bounding boxes, OCR-extracted text, and LLM-verified outputs. The users can review the system’s confidence scores, inspect cropped regions, and receive intelligent feedback or correction suggestions directly from the LLM agent. This step of reasoning is facilitated through a LangChain interface enabling real-time prompting and post-processing checking of outcomes. This interactive platform streamlines the analysis process, making scale bar validation accessible and efficient for non-technical users. 

This proposed LLM agent has the object detection and information extraction context fed into the backend, allowing it to reason using structured data like pixel length, OCR value, unit, confidences, etc. before generating its answer from the SEM image analysis pipeline. Therefore, in comparison to other general LLMs like Claude or GPT, it acts as domain assistant designed for SEM analysis. Figure \ref{fig:combined}a shows the architecture flow of the LLM evaluation process. In order, to evaluate the performance of this proposed LLM agent, it is compared with other raw general-purpose multi-modal LLMs including: 1) Llama 4 Maverick and 2) Llama 4 Scout , as seen in Figure \ref{fig:combined}b. For the LLM evaluation process, a benchmark of twenty questions was generated to analyze the performance of the Integrated LLM-Based Agent \citep{ref28,ref29,ref30} (given in Appendix B). These questions generally are related to the morphology and structures of the materials, the scale dimensions, the material composition and any imaging parameters. A total of 127 SEM images are fed to the LLM models with 20 benchmark questions and their outputs are recorded for each SEM image. Then, these outputs are passed to an LLM acting as an expert. The prompt for the evaluation is given (See Appendix B). The scores are later aggregated per image and given in Appendix B, Table B1. The evaluation metrics include a) accuracy of predicting the scale value and unit, b) detection quality, c) OCR reliability, d) reasoning quality and e) practicality for each SEM image. The rating (1-5) of the Integrated LLM-Based Agent per question of the SEM images is also given in Appendix B, Table B1. The integrated LLM-Based agent has a better accuracy of 70\% and a reliable OCR confidence with high quality reasoning in SEM related images but has inconsistent latency. This results show that the responses demonstrate a strong fundamental knowledge of the material characterization. Overall, the responses reflect strong technical SEM image analysis particularly in defect detection, material property correlations but shows certain inconsistencies in structural descriptions. The Integrated LLM-Based Agent responses are generally strong with high accuracy and completeness for the benchmark questions. The Maverick model excels in speed and provides a good general SEM image analysis but cannot detect or measure the scale bars. Similarly, the Scout model performs better in speed and gives a clean output but produces scientifically invalid results too frequently. 
\begin{center}
% First part
\noindent
\includegraphics[width=0.6\textwidth]{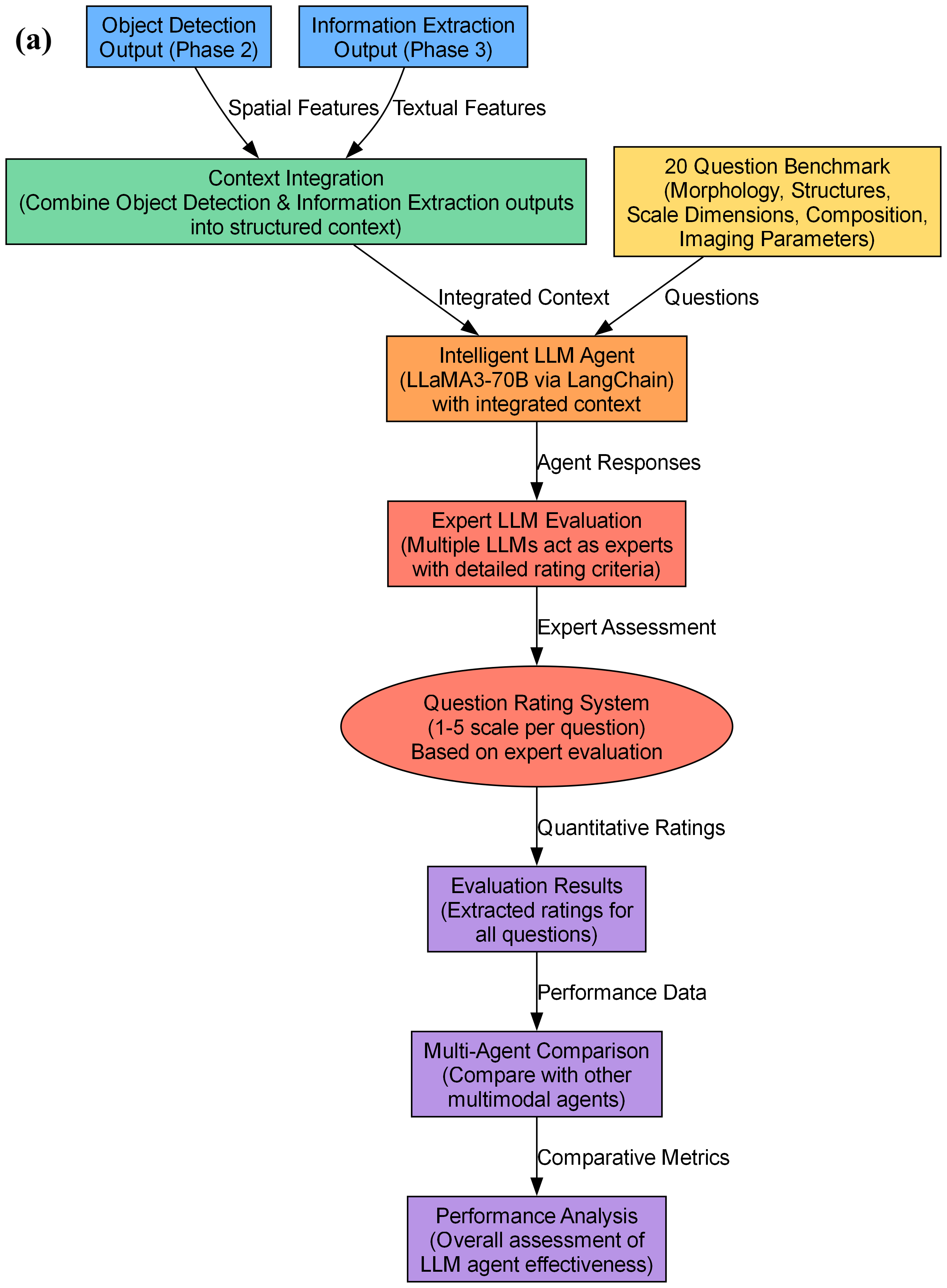}
\par\vspace{0.3cm}
\includegraphics[width=0.9\textwidth]{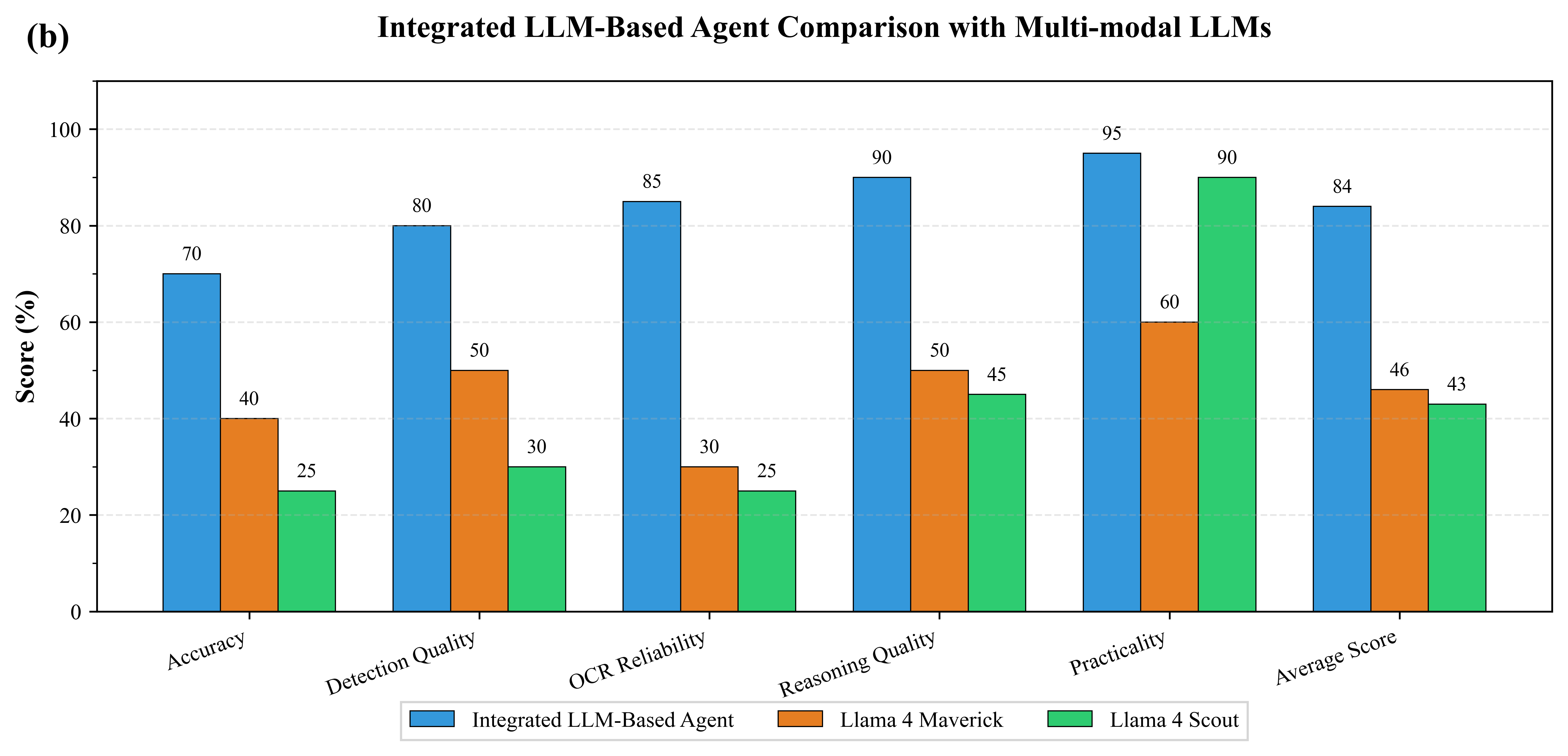}
\par\vspace{0.3cm}
\captionof{figure}{(a) Flowchart outlining the process of the LLM agent evaluation. (b) Integrated LLM-Based Agent Comparison with other multi-modal General Purpose LLMs.}
\label{fig:combined}
\end{center}
This makes the integrated LLM-Based agent a much reliable, well-reasoned output for real-world applications that need better accuracy and trustworthiness more than speed.
\section{Conclusion and Future Work}\label{sec4:conclusion}
In this paper, the auto-detection of scale bars in SEM images is explored with a multi-modal and automated object detection and information extraction system that involves an Auto-DG module to create a heterogeneous and representative dataset. This Auto-DG module generates synthetic scale bars with random lengths, thicknesses, colors, units, and font styles for scanning electron microscope images. Each image was automatically labeled with bounding box and text metadata to be compatible with object detection and OCR training pipelines. Next, deep learning techniques are used to automate the detection and extraction of scale bars demonstrating a strong performance in object detection and accurate localization with a precision of 100\%, recall of 95.8\%. The OCR module utilized a hybrid strategy of DenseNet-based feature learning trained to project visual features of cropped scale bar regions to text representation. The OCR model emits both the predicted scale value and unit, together with a confidence score for every prediction offering a 89\% precision, 65\% recall, and a 75\% F1 score on the Auto-DG dataset. Lastly, the LLM agent, driven by a domain-adapted variant of LLaMA3, was incorporated to reason over OCR outputs in context, verify the identified scale values, and disambiguate based on user-specified prompts offering a 70\% accuracy.

Therefore, this LLM based SEM OCR proposed system has demonstrated an exceptional accuracy and efficiency, significantly reducing the manual labor and potential for error associated with the traditional methods of scale bar detection. With extensive testing on various datasets, the proposed object detection and information extraction model has proven its robustness and reliability across a wide array of SEM images with different positions and complexities of the scale bars. This capability ensures that researchers can rely on the model to provide precise measurements crucial for detailed microscopic analysis. Future work could focus on developing a customized model specifically tailored to cope with the unique challenges posed by colored and complex backgrounds. This might involve augmenting the training dataset with a more diverse set of images, including those with colored backgrounds, to better generalize the model’s capabilities. Moreover, we shall focus on refining recall through adaptive thresholds and expanding unit recognition to include less common measurements. These results establish the combined approach as a state-of-the-art tool for enhancing accuracy in microscopic scale bar detection.

The implications of this work extends beyond simple scale bar detection, opening up new possibilities for automation in scientific imaging. This can potentially enhance the productivity of researchers and provide more consistent and reliable data for scientific studies. Overall, the development of the Integrated LLM-Based Agent represents a significant advancement in the field of microscopic image analysis, offering a powerful tool that meets the growing demands for automation and precision in scientific research. 

\bibliographystyle{unsrtnat}
\bibliography{references}  %%% Uncomment this line and comment out the ``thebibliography'' section below to use the external .bib file (using bibtex) .

%%% Uncomment this section and comment out the \bibliography{references} line above to use inline references.
% \begin{thebibliography}{1}

% 	\bibitem{kour2014real}
% 	George Kour and Raid Saabne.
% 	\newblock Real-time segmentation of on-line handwritten arabic script.
% 	\newblock In {\em Frontiers in Handwriting Recognition (ICFHR), 2014 14th
% 			International Conference on}, pages 417--422. IEEE, 2014.

% 	\bibitem{kour2014fast}
% 	George Kour and Raid Saabne.
% 	\newblock Fast classification of handwritten on-line arabic characters.
% 	\newblock In {\em Soft Computing and Pattern Recognition (SoCPaR), 2014 6th
% 			International Conference of}, pages 312--318. IEEE, 2014.

% 	\bibitem{hadash2018estimate}
% 	Guy Hadash, Einat Kermany, Boaz Carmeli, Ofer Lavi, George Kour, and Alon
% 	Jacovi.
% 	\newblock Estimate and replace: A novel approach to integrating deep neural
% 	networks with existing applications.
% 	\newblock {\em arXiv preprint arXiv:1804.09028}, 2018.

% \end{thebibliography}

\end{document}